\documentclass[sigconf]{acmart}
\usepackage{libertine}
\usepackage{graphicx}
\usepackage{subfigure}

\usepackage{multirow}

\usepackage[medium,compact]{titlesec}

\newcommand{\tabincell}[2]{
	\begin{tabular}{@{}#1@{}}#2\end{tabular}
}

\AtBeginDocument{%
  \providecommand\BibTeX{{%
    \normalfont B\kern-0.5em{\scshape i\kern-0.25em b}\kern-0.8em\TeX}}}

\copyrightyear{2022} 
\acmYear{2022} 
\setcopyright{acmcopyright}\acmConference[MMAsia '22]{ACM Multimedia Asia}{December 13--16, 2022}
{Tokyo, Japan}
\acmBooktitle{ACM Multimedia Asia (MMAsia '22), December 13--16, 2022, Tokyo, Japan}
\acmPrice{15.00}
\acmDOI{10.1145/3551626.3564968}
\acmISBN{978-1-4503-9478-9/22/12}

\acmSubmissionID{59}

\begin{document}

\title{Towards High Performance One-Stage Human Pose Estimation}


\author{Ling Li$^{1,2}$, Lin Zhao$^{1,2\dag}$, Linhao Xu$^{1}$, Jie Xu$^{1}$}
\affiliation{%
  \institution{{$_{1}$} PCA Lab, Key Lab of Intelligent Perception and Systems for High-Dimensional Information of Ministry of Education, and Jiangsu Key Lab of Image and Video Understanding for Social Security, Nanjing University of Science and Technology}
}
\affiliation{%
  \institution{{$_{2}$} State Key Laboratory of Integrated Services Networks (Xidian University)}
}
\email{{lingl, linzhao, linh, jiexu}@njust.edu.cn}
\renewcommand{\shortauthors}{Ling Li, et al.}



\begin{abstract}
Making top-down human pose estimation method present both good performance and high efficiency is appealing. Mask RCNN can largely improve the efficiency by conducting person detection and pose estimation in a single framework, as the features provided by the backbone are able to be shared by the two tasks. However, the performance is not as good as traditional two-stage methods. In this paper, we aim to largely advance the human pose estimation results of Mask-RCNN and still keep the efficiency. Specifically, we make improvements on the whole process of pose estimation, which contains feature extraction and keypoint detection. The part of feature extraction is ensured to get enough and valuable information of pose. Then, we introduce a Global Context Module into the keypoints detection branch to enlarge the receptive field, as it is crucial to successful human pose estimation. On the COCO val2017 set, our model using the ResNet-50 backbone achieves an AP of 68.1, which is 2.6 higher than Mask RCNN (AP of 65.5). Compared to the classic two-stage top-down method SimpleBaseline, our model largely narrows the performance gap (68.1 $AP^{kp}$ vs. 68.9 $AP^{kp}$) with a much faster inference speed (77 ms vs. 168 ms), demonstrating the effectiveness of the proposed method. Code is available at: https://github.com/lingl\_space/maskrcnn\_keypoint\_refined.
\end{abstract}




\maketitle

\section{Introduction}
Multi-person pose estimation has always been a fundamental and challenging task in computer vision. It has many complex downstream applications, such as human parsing \cite{humanparsing,jointparsing,reidentificationfeature,partparsing}, action recognition \cite{semanticaligned,glimpse,3dpose}, human-computer interaction \cite{humanrobot1,humanrobot2}, video surveillance \cite{video1,video2,video3,video4}, and animation generation \cite{animation1,animation2}.

\begin{figure}[tbp]
	\centering
	\includegraphics[width=4.5cm]{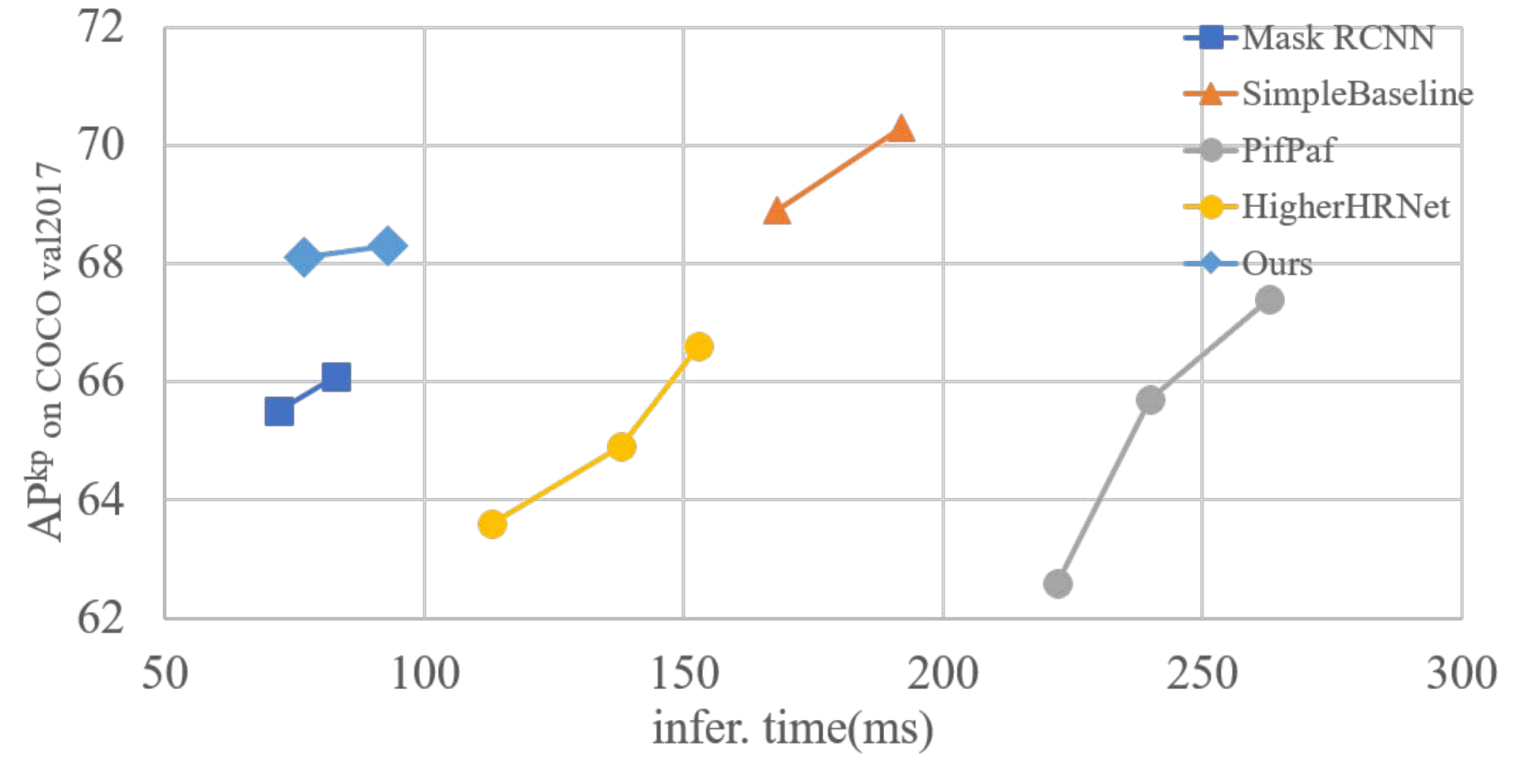}
	\caption{The trade-off between speed and accuracy. Inference time is measured on a single TITAN V GPU.}
	\label{infer}
\end{figure}

With the continuous development of deep convolutional neural networks (CNN), the task of human pose estimation has achieved remarkable progress. The main ideas are roughly divided into two groups: top-down methods \cite{grmi,rmpe,simplebaseline,hrnet} and bottom-up methods \cite{deepcut,openpose,ae,rethink1}. Although bottom-up methods have better real-time performance, it is difficult to achieve the same high performance as top-down methods, due to the diversity of human instances scales. Generally, top-down methods obtain more accurate keypoint localization based on the detected bounding boxes. But because throughout the process features cannot be shared by two independent steps, the inference time can be much longer.

Mask RCNN \cite{maskrcnn} proposes the technique of RoIAlign to replace the traditional RoIPooling, which solves the problem of misalignment between CNN features and the input image. With this technique, Mask RCNN successfully integrates keypoint detection into the framework of person instance detection. Thus, the efficiency can be largely improved, since the keypoint detection branch can directly use the features from the backbone. However, Mask RCNN's performance is far behind traditional two-stage top-down methods like SimpleBaseline \cite{simplebaseline}. We make a thorough investigation on the keypoint detection process of Mask RCNN, and aim to largely refine the results but still keep the efficiency (as shown in Fig. \ref{infer}).

First of all, due to the inaccuracy of the candidate boxes, the contextual information around some keypoints can easily exceed the box and be discarded, it may affect the keypoint detection. Moreover, the Feature Pyramid Network (FPN) \cite{fpn} can make the object detection results robust to the various size. Yet pose estimation is a pixel-level task, which requires finer feature granularity, and high resolution is important to get accurate results, as demonstrated in HRNet \cite{hrnet}. Thus, it may not be proper to use the same feature selection strategy as object detection. Third, the perception of human body structure is very crucial to pose estimation. It is necessary to expand the receptive field to provide more contextual information. However, the original design in Mask RCNN focuses more on local information mining since the network depth is limited. 

\begin{figure}[tb]
	\centering
	\includegraphics[width=4.8cm]{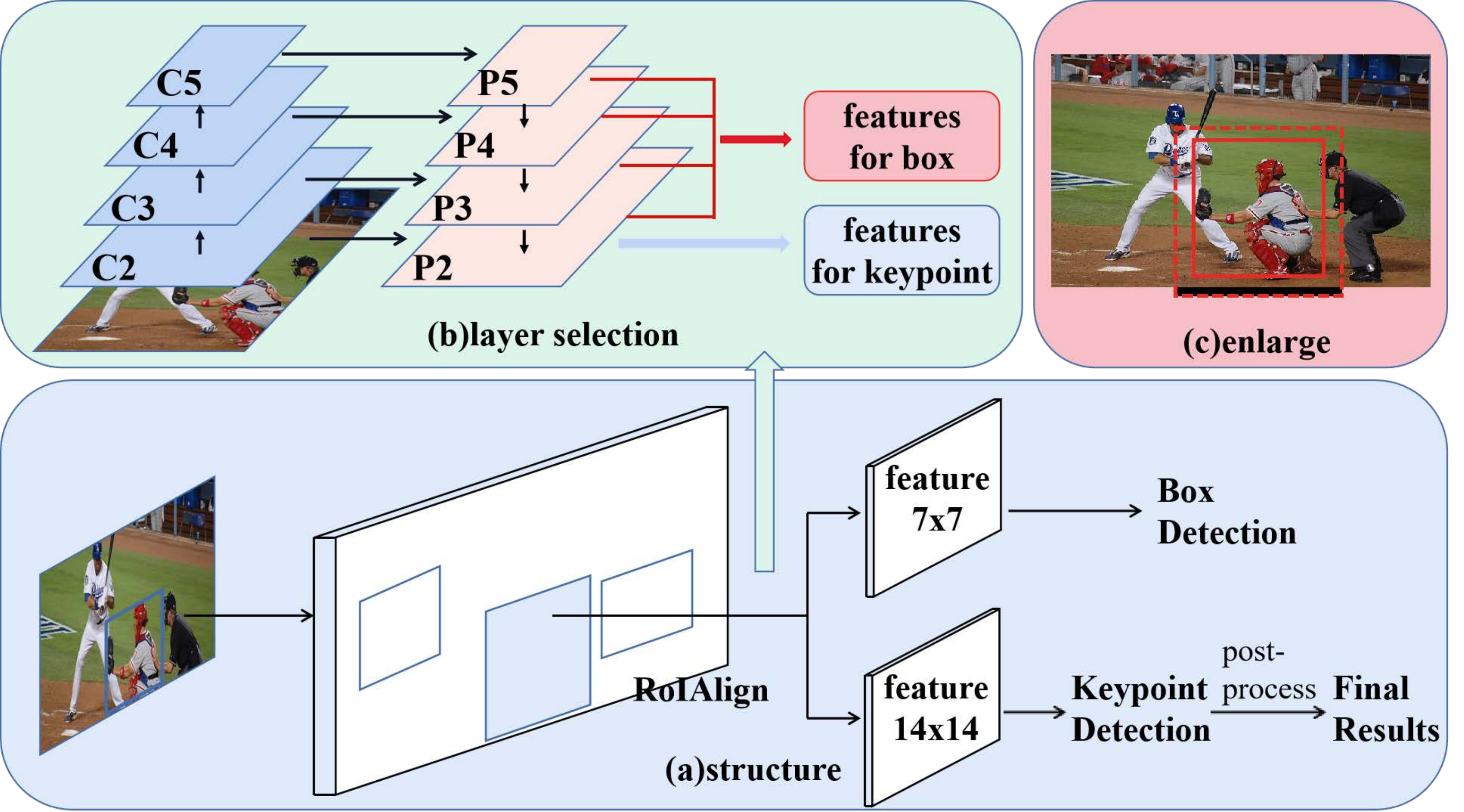}
	\caption{The refinements based on Mask RCNN. (a) describes the complete process. (b) explains different strategies for FPN layer selection. The strategy of enlarging box is shown in (c).}
	\label{structure}
\end{figure}

In this paper, we demonstrate that a comprehensive  solution (as shown in Fig. \ref{structure}) of all the above aspects is able to improve the pose estimation performance of Mask RCNN greatly, and still keep the superior efficiency over traditional two-stage top-down methods.

\section{Related Work}

\subsection{Traditional Two-Stage Methods}

\paragraph{Top-down Methods.} The top-down strategy is composed of two phases: detecting all human instances for image cropping and then acquiring pose by single person keypoint detector. Well-known methods include Hourglass \cite{hourglass}, Simplebaseline \cite{simplebaseline}, HRNet \cite{hrnet}, PPNet \cite{ppnet} and so on. Since top-down methods normalize the cropped image size, it is more robust to human instances of different scales. And it is able to obtain relatively high-resolution features, which is helpful to keypoint location. But the inference speed is slow due to the inability to share computation and features with the object detector. This main drawback makes the inference time increase linearly with the number of human instances in one image.

\paragraph{Bottom-up Methods.} 
Unlike top-down methods, bottom-up methods predict all possible keypoints, and assign them to corresponding human instances \cite{2dpose}. There are various association algorithms, such as dynamic programming \cite{openpose}, tag matching \cite{ae}, greedy decoding algorithm \cite{personlab}. Their computational complexity is not related to the number of human instances and can compute all features once. However, it is difficult to obtain accurate localization due to the sensitivity to instance scale changes, and the matching during inference is an NP-hard problem with high complexity.

\subsection{One-Stage Methods}

Two-stage methods, especially those using top-down strategies, obtain good performances in accuracy, but have low inference efficiency. So one-stage methods are proposed, which combine human detection with pose estimation. A one-stage solution for pose estimation, the SPM model \cite{spm}, is presented by Nie et al. It uses a root joint to represent the human body position and then uses offsets to represent keypoints. While ensuring the speed advantage, this scheme achieves performance on par with the two-stage bottom-up methods. Recently, Geng et al. \cite{dekr} present a competitive one-stage method, DEKR, that employs a novel pose-specific neural network to solve keypoint regression. Based on the foundation of anchor-free object detector FCOS \cite{fcos}, Mao et al. propose a fully convolutional multi-person pose estimation network, FCPose \cite{fcpose}, which uses dynamic instance-aware convolutions for pose estimation. 

\subsection{Our approach}

Mask RCNN \cite{maskrcnn} can also be viewed as a one-stage method. 
But different from one-stage methods mentioned above, it also uses top-down strategy as the bounding boxes are needed for separating different persons. In this paper, we aim to thoroughly refine the process of pose estimation based on the Mask RCNN framework, achieving the performance close to two-stage top-down methods, and keeping the efficiency of one-stage methods.

\section{Method}

\begin{figure}[tb]
    \centering
	\subfigure[Keypoints miss detection situation]{
		\centering
		\includegraphics[width=3.9cm]{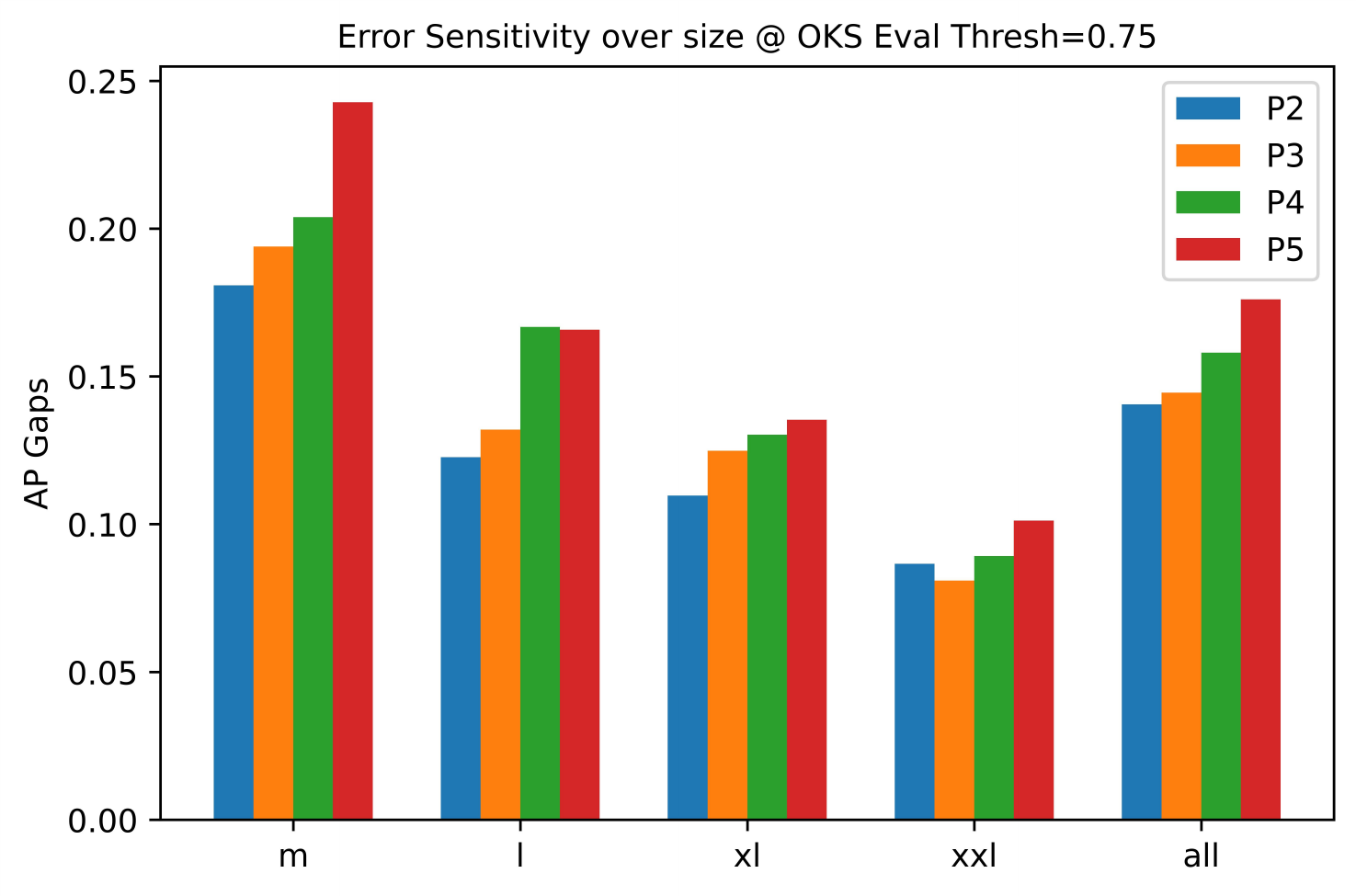}
		\label{a}
	}\quad
	\subfigure[The information needed for keypoints]{
		\centering
		\includegraphics[width=3.9cm]{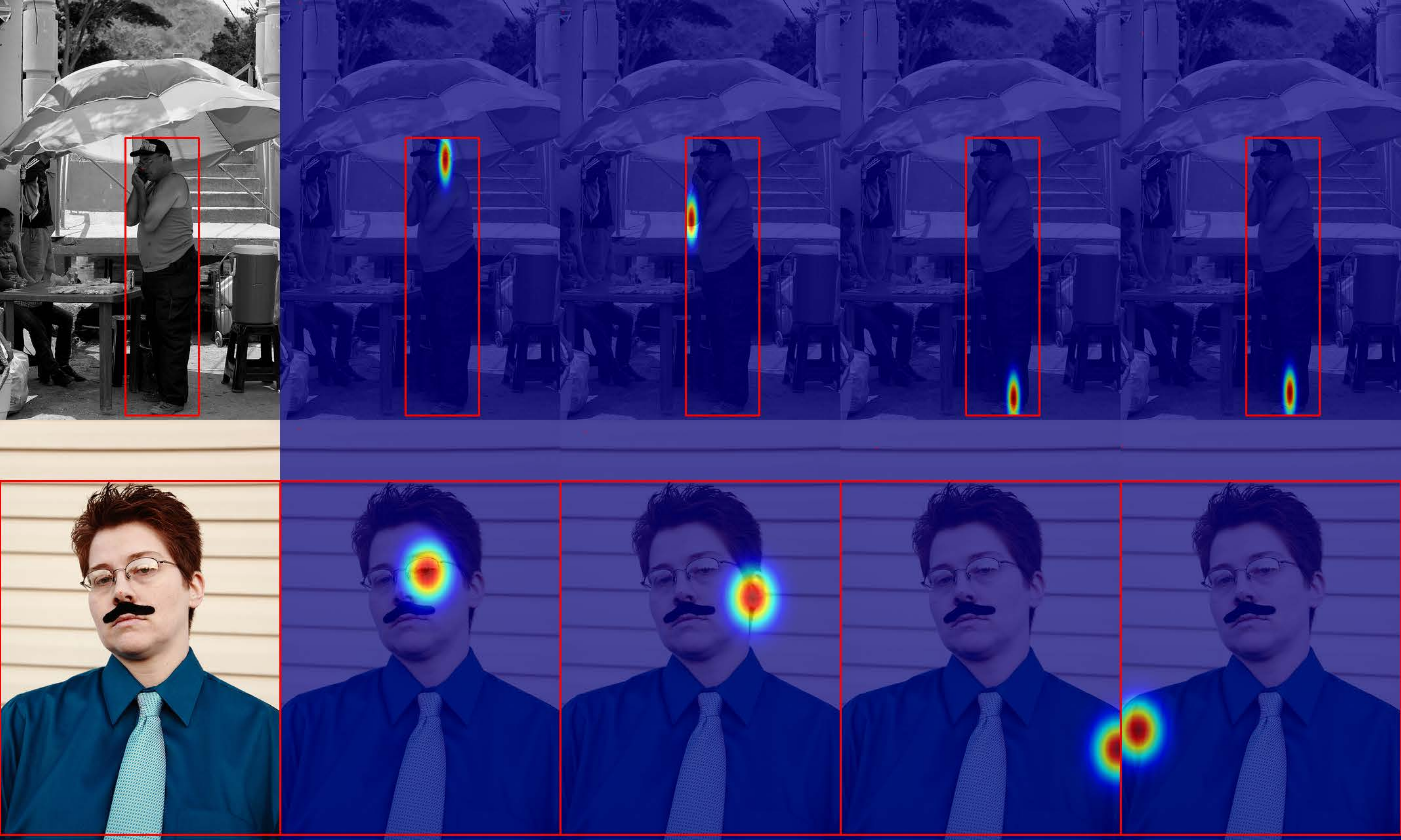}
		\label{b}
	}
	\quad
	\centering
	
	\caption{(a) shows keypoint missed detections resulting from training with either FPN layer. And (b) shows using tight bounding boxes tends to discard important features.
	}
	\label{fpnimpact}
\end{figure}

\subsection{Feature extraction during RoIAlign}

\subsubsection{The selection of feature layers.} 

The selection of output feature layers in FPN \cite{fpn} has a certain influence on the subsequent tasks. We can regard keypoints as small-scale objects, thus requiring strong spatial information for detection. Typically, RoIAlign uses box regions to select in layers P2 to P5 of FPN to extract features. However, the low resolution of high-level features can easily lead to missed detection of keypoints. As shown in Fig. \ref{fpnimpact}, we analyzed the miss situation when taking features on each layer, and found that small-scale persons have the most serious misses when taking features on $P_{4}$ and $P_{5}$.
Here, we propose to use only the $P_{2}$ feature layer for keypoint detection, which not only has high resolution and more precise spatial information, but also retains strong semantic information due to the top-down feature fusion process in FPN.

\subsubsection{The strategy for bounding box enlarging.} 

The size of bounding box determines the amount of pose-related information the keypoint detection branch can obtain. We need to choose the best-fit box size to provide sufficient and effective features.

We use heatmaps to represent sensitive feature regions during keypoint detection, as shown in Fig. \ref{fpnimpact}. The heat values represent the importance of the surrounding contextual features. We map the area back to the original image and use red boxes to represent the box detection result. As can be seen, due to the given proposal box is not always very accurate and may surround the person instance tightly, RoIAlign is very probable to discard the features important for keypoint detection, especially for the keypoints adjacent to the edge of detection boxes. It can be seen from the visualization results in Fig. \ref{fpnimpact} that shoulders, elbows, wrists, and ankles are more prone to be out-of-bounds. The loss of keypoint-sensitive features is irreparable, which will have a great impact on the results. To solve this problem, we experimentally determine that a magnification of 1.3$\times$ is appropriate. Meanwhile, we fill the area beyond image with black backgrounds to prevent human body center deviation.

\begin{figure}[tb]
	\centering
	\subfigure[The refined keypoint detection branch]{
		\centering
		\includegraphics[width=4.5cm,height=2.8cm]{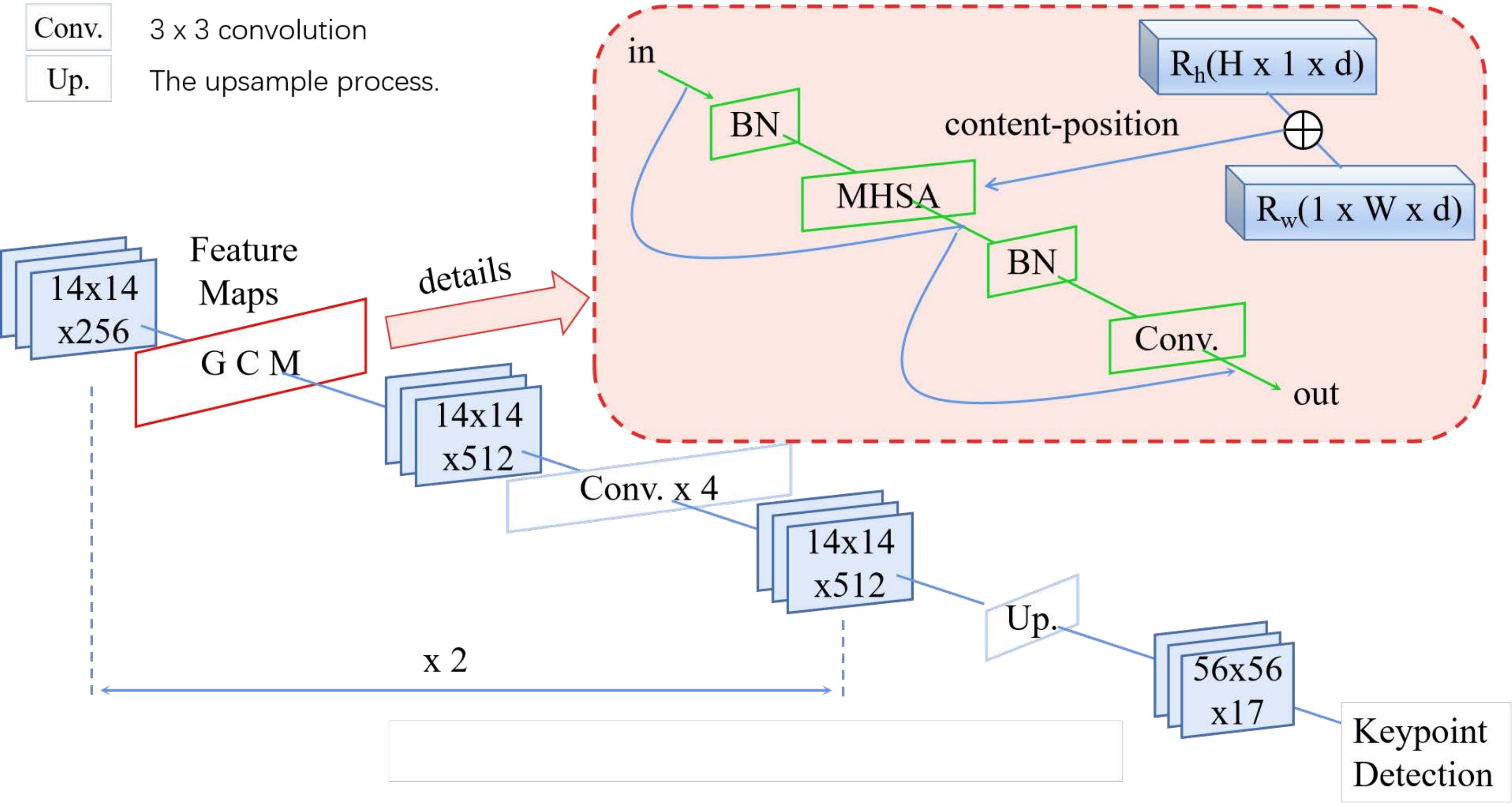}
	}\quad
	\subfigure[Fusion structure]{
		\centering
		\includegraphics[width=2.7cm,height=2.8cm]{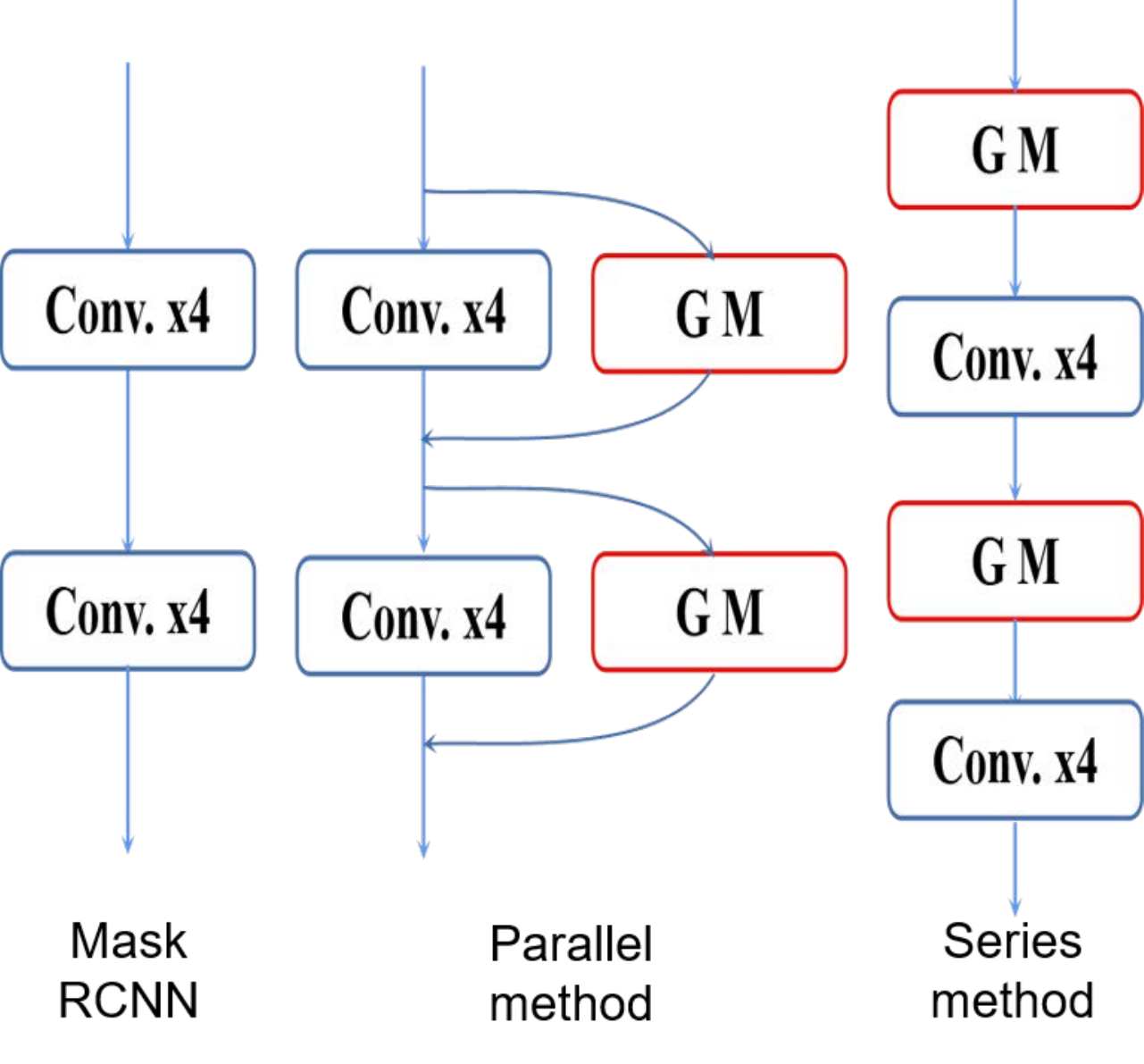}
		\label{b}
	}
	\quad
	\caption{
	Our keypoint branch and two fusion structures. 
	}
	\label{network}
\end{figure}

\subsection{The Global Context Module
}

A successful pose estimation model needs to incorporate contextual information in a sufficiently large receptive field to learn the complex relationship of body parts. Mask RCNN uses 8 convolutions for detection with a receptive field of 17, which is not enough.

\subsubsection{The Structure of Module.}

Blindly stacking convolutions to enlarge the receptive field will greatly increase the model size and reduce the inference speed. Inspired by the successful applications of transformer models like ViT \cite{vit}, we utilize the multi-head self-attention (MHSA) to solve the problem.

We refer to the added part as Global Context Module (GCM). As shown in the Fig. \ref{network}, it is composed of two parts. One is a multi-head self-attention module referring to the BoTNet \cite{botnet}, which is used to extract global information. The other is a 3$\times$3 convolution kernel to perform preliminary integration of global features while aligning the channels. In addition, the residual structure is added. By using this module, we can provide the model with global information, and the performance of keypoint detection is able to be improved.

\subsubsection{Fusion Structure.}

After acquiring global features, we fuse them with local features in a series connection. As shown in the Fig. \ref{network} (a), first, we use the GCM to extract global context, and then use 4 convolutions to further process the features. This process is repeated twice to obtain the information for final pose estimation.

\section{Experiment results}

\begin{figure}[tbp]
	\centering
	\subfigure[Enlarging boxes of different sizes]{
		\centering
		\includegraphics[width=3.5cm]{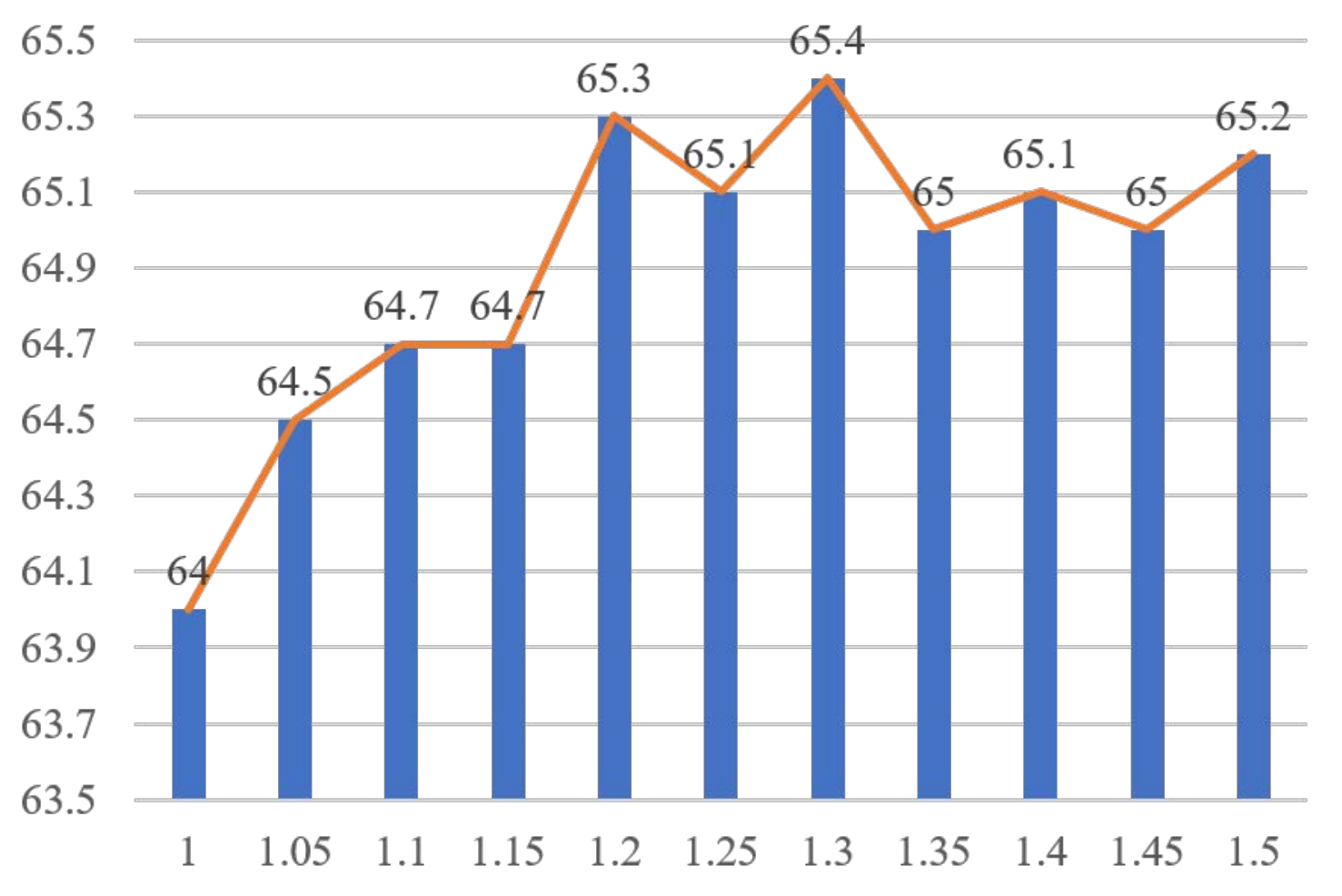}
		\label{a}
	}\quad
	\subfigure[Using different layers of FPN]{
		\centering
		\includegraphics[width=3.5cm]{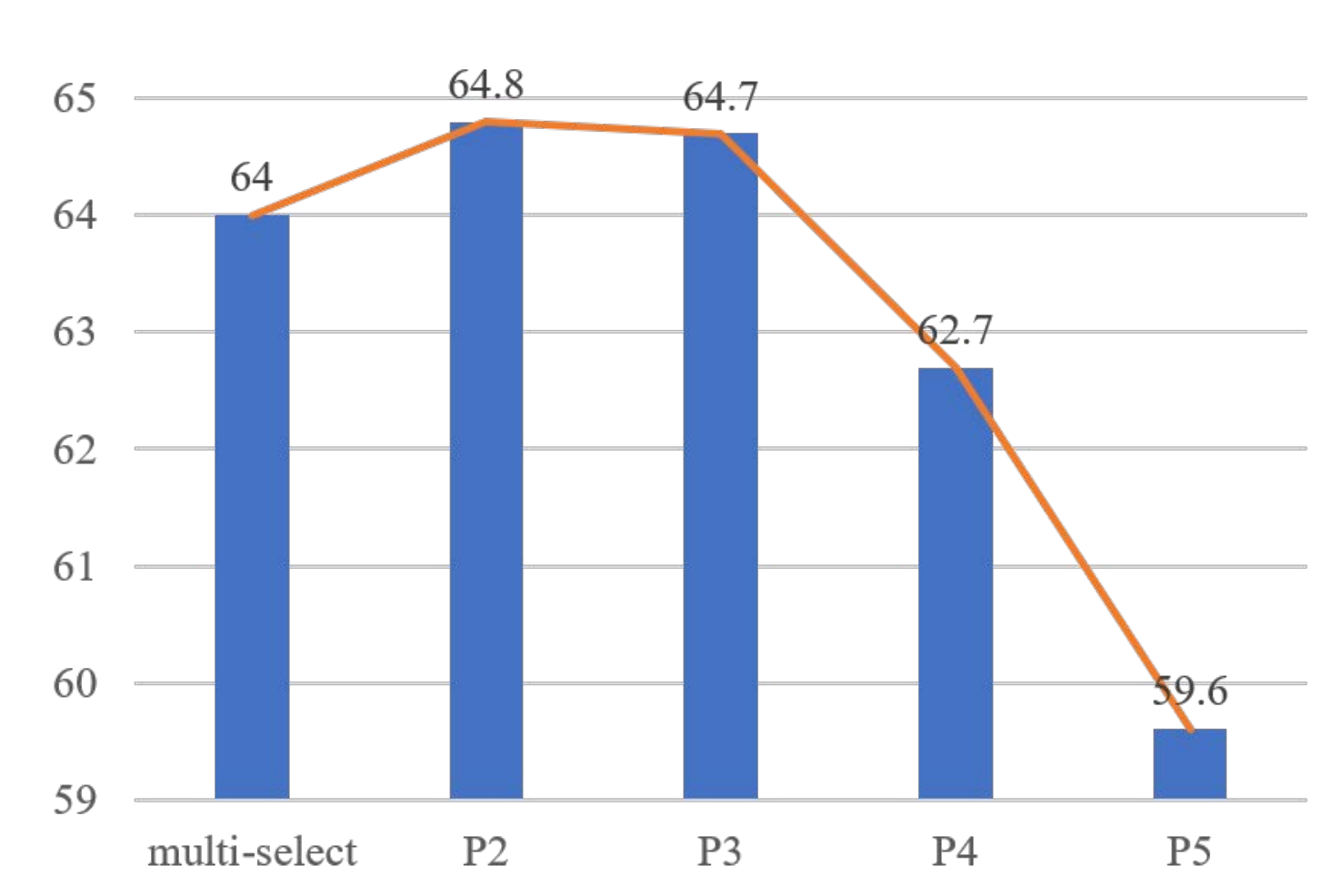}
		\label{b}
	}
	\quad
	\centering
	\caption{Ablation study of Different Feature Extraction Strategies in RoIAlign. The 1.0x magnification and multi-select experiments refer to the strategy in the Mask RCNN.
	}
	\label{dataprocessing}
\end{figure}

\subsection{Dataset}

The COCO dataset \cite{coco} contains over 250K person instances, each of which contains 17 annotated keypoints. We train our model on the COCO train2017, which contains 57K images and 150K human instances. The model is then validated on COCO val2017 and COCO test-dev2017, with 5K images and 20K images respectively.

\subsection{Implementation details}

We use Detectron2 \cite{detectron2} for the model implementation. The model is trained using a single TITAN V GPU with Stochastic Gradient Descent. We initialize with a backbone model pretrained on the ImageNet classification task \cite{init1}. The base learning rate is initially set to 0.0025, and is reduced to a tenth of the original at 480K and 640K iterations in the 1$\times$ training schedule (i.e., 720K iterations). In terms of data augmentation, we adjust the longer side of images to be less than or equal to 1333 and the shorter side to be between 640 and 800. Random flipping is used to enhance the learning ability.

All inference processes are measured on a single TITAN V GPU. The shapes of the images are adjusted to have a length of 800 on the short side, or no more than 1333 on the long side. Most importantly considering the high efficiency maintenance, we do not use flip operation and multi-scale testing strategy to produce better results.

\subsection{Ablation Experiments}

In this section, we investigate the performance of our improved model in 1x training schedule with ResNet-50 as the backbone. 

\subsubsection{Comparison of Different Feature Extraction Strategies in RoIAlign.}

We set the magnification of bounding boxes at intervals of 0.05 in the range of 1 to 1.5 for experiments. Fig. \ref{dataprocessing} (a) shows the experimental results. The AP increases almost linearly from 1.0 to 1.3 due to the increased information around keypoints. However, AP decreases slowly when the magnification exceeds 1.3 due to the introduction of a lot of useless background features. Thus, we select to enlarge bounding box by 1.3x to obtain the best performance. 

\begin{table*}[thb]\tiny
	\centering
	\caption{
	Comparison of experimental results. Mask RCNN uses the implementation on Detectron2, and $-$ represents testing with the boxes we measured, and turning off the flip test. The rescoring network is not used when DEKR tests.
	}
	\begin{tabular}{c|c|c|c|c|cc|cc|c|cc|cc|}
	    \toprule
	    & & & & \multicolumn{5}{c|}{COCO val2017} & \multicolumn{5}{c|}{COCO test-dev 2017}\\
		&Method&backbone&\tabincell{c}{infer.\\(ms)}&$AP^{kp}$&$AP^{kp}_{50}$&$AP^{kp}_{75}$&$AP^{kp}_M$&$AP^{kp}_L$&$AP^{kp}$&$AP^{kp}_{50}$&$AP^{kp}_{75}$&$AP^{kp}_M$&$AP^{kp}_L$\\
		\midrule
		\multirow{4}{*}{\rotatebox{90}{Top-down}}&
		 G-RMI \cite{grmi}&ResNet-101&-&-&-&-&-&-&64.9&85.5&71.3&62.3&70.0\\
        &SimpleBaseline- \cite{simplebaseline}&ResNet-50&168&68.9&88.2&76.5&65.5&75.2&67.7&89.0&75.2&64.4&73.6\\
        &SimpleBaseline- \cite{simplebaseline}&ResNet-101&192&70.3&89.0&78.0&67.0&76.4&69.0&89.9&77.1&65.9&74.7\\
        &HRNet- \cite{hrnet}&HRNet-W32&-&73.4&90.0&80.6&69.7&79.9&71.5&90.5&79.6&68.1&77.4\\\hline
        \multirow{4}{*}{\rotatebox{90}{Bottom-up}}&SimplePose \cite{simplepose}&IMHN&-&66.1&85.9&71.6&59.8&76.2&68.5&86.7&74.9&66.4&71.9\\
        &PersonLab \cite{personlab}&ResNet-152&-&66.5&86.2&71.9&62.3&73.2&66.5&88.0&72.6&62.4&72.3\\
        & PifPaf \cite{pifpaf}&ResNet-152&263&67.4&-&-&-&-&66.7&87.8&73.6&62.4&72.9\\
        &HigherHRNet \cite{higherhrnet}&HRNet-W32&128&67.1&86.2&73.0&61.5&76.1&64.7&86.9&71.0&60.2&71.2\\\hline
        \multirow{6}{*}{\rotatebox{90}{One-Stage}}&CenterNet \cite{centernet}&Hourglass&147&64.0&85.6&70.2&59.4&72.1&63.0&86.8&69.6&58.9&70.4\\
        &Mask RCNN \cite{maskrcnn}&ResNet-50&72&65.5&87.2&71.1&61.3&73.4&63.1&87.3&68.7&57.8&71.4\\
        &Mask RCNN \cite{maskrcnn}&ResNet-101&83&66.1&87.4&72.0&61.5&74.4&-&-&-&-&-\\
        &FCPose \cite{fcpose}&ResNet-50&68&-&-&-&-&-&64.3&87.3&71.0&61.6&70.5\\
        &FCPose \cite{fcpose}&ResNet-101&93&-&-&-&-&-&65.6&87.9&72.6&62.1&72.3\\
        &DEKR \cite{dekr}&HRNet-W32&63&67.2&86.3&73.8&61.7&77.1&66.6&87.6&73.5&61.2&75.6\\
        &SPM \cite{spm}&Hourglass&-&-&-&-&-&-&66.9&88.5&72.9&62.6&73.1\\\hline
        &\textbf{Our method}&\textbf{ResNet-50}&\textbf{77}&\textbf{68.1}&\textbf{88.0}&\textbf{74.5}&\textbf{63.7}&\textbf{76.2}&\textbf{66.4}&\textbf{88.4}&\textbf{73.1}&\textbf{62.2}&\textbf{73.9}\\
        &\textbf{Our method}&\textbf{ResNet-101}&\textbf{90}&\textbf{68.3}&\textbf{88.0}&\textbf{74.8}&\textbf{63.8}&\textbf{76.6}&\textbf{67.1}&\textbf{89.0}&\textbf{74.0}&\textbf{63.0}&\textbf{74.7}\\
        \bottomrule
	\end{tabular}
	\label{allresults}
\end{table*}

\begin{table}[thb]\tiny
	\centering
	\caption{Comparative experiments on network structures. 
	}
	\begin{tabular}{c|c|c|c|c|c}
        \toprule
		\textbf{Method}&$AP^{kp}$&$AP^{kp}_{50}$&$AP^{kp}_{75}$&$AP^{kp}_M$&$AP^{kp}_L$\\
		\midrule
		Mask RCNN & 64.0 & 86.0 & 69.7 & 59.6 & 72.1\\\hline
		$GCM_{parallel}$ & 64.6 & 86.2 & 70.2 & 60.0 & 72.9\\\hline
		\textbf{$GCM_{series}$} & \textbf{65.2} & 86.8 & \textbf{71.1} & \textbf{60.7} & \textbf{73.4}\\\hline
	\end{tabular}
	\label{atrous}
\end{table}

As for the selection of feature levels, we also conduct separate experiments for each level. The results are shown in Fig. \ref{dataprocessing} (b). Mask RCNN uses the sizes of bounding boxes to select different feature layers from FPN, which only obtain 64 AP. Using the $P_{2}$ layer alone to acquire features can achieve the best performance. The experiments using the $P_{4}$ and $P_{5}$ layer alone show severe AP drops, by 1.3 and 4.4 AP respectively. Referring to the results of error analysis \cite{benchmark} in Fig. \ref{fpnimpact}, we conclude that keypoint missed detection is particularly serious when taking features at the $P_{4}$ or $P_{5}$ layer.

Based on the above experiments, we enlarge bounding boxes by 1.3x and select the $P_{2}$ layer to get pose-related features. The AP is improved by 2.1 from 64.0 to 66.1 using the ResNet-50 backbone.

\subsubsection{The Design of the Keypoint Branch.} 

We present two fusion methods in Fig. \ref{network}. The results given in Table \ref{atrous} show that the series fusion method leads to better performance. One possible explanation is that obtaining some global context information is more beneficial to subsequent local feature extraction.

\subsubsection{Combination of All Techniques.} 

Finally, we combine all the refinement strategies used above. The final AP can be improved to 66.8, outperforming original Mask RCNN by 2.8 AP, which demonstrates the effectiveness of the proposed method.

\subsection{Comparison with the State-of-the-art Methods}

In this section, we compare our model with other methods. Here our model and Mask RCNN are trained using 3$\times$ training schedule. The results of other methods are obtained using their public trained models or from the corresponding papers. We report the results of our model and other state-of-the-art methods in Table \ref{allresults}.

\subsubsection{Comparison with Two-Stage Methods.} 

We mainly compare our method with SimpleBaseline \cite{simplebaseline}.  Because SimpleBaseline is the most classic and its structure is as simple as ours. For a fair comparison, we turn off the flip test, and the person instance detection boxes from our models (Person AP of 55.4) are used. The performance of our method still lags behind 0.8 AP and 2.0 AP under ResNet-50 and ResNet-101 backbone respectively. However considering the raw performance gap between Mask RCNN and SimpleBaseline, this has been narrowed largely. Moreover, using the ResNet-50 backbone, our method averagely takes 77 ms to detect keypoints from one image, which is much faster than SimpleBaseline's 168 ms. Under the ResNet-101 backbone, our method is still more than 2x faster. Because high efficiency is as important as performance in real applications, our method may be a better solution. Compared to other two-stage top-down methods like the famous HRNet \cite{hrnet}, the performance of our method is not competitive, however, their inference time is much larger than ours.

Furthermore, we compare our model with bottom-up methods. Our model using the ResNet-50 backbone can achieve higher detection performance than HigherHRNet using the HRNet-W32 backbone. Compared with most bottom-up methods, we have advantages both in inference time and detection performance.

\subsubsection{Comparison with One-Stage Methods.} 

Compared with the benchmark model Mask RCNN, our model can be closer to the state-of-the-art top-down methods while maintaining real-time performance. Meanwhile, compared with other one-stage methods, our model also obtains better results. Our model with the ResNet-50 backbone outperforms CenterNet \cite{centernet} 4.1 AP with about half the inference time. And when we turn off the rescoring network in DEKR \cite{dekr} for a fair comparison, our model achieves better performance with a small backbone ( 68.1 AP vs. 67.2 AP).

\subsubsection{Results on COCO test-dev}

Our model is able to achieve better or comparable AP performance on the COCO test-dev2017 compared to most bottom-up and one-stage methods. Our model using the ResNet-101 backbone achieves 67.1 AP. Compared to the most recent FCPose \cite{fcpose} with the same backbone, our model receives 2.1 AP (ResNet-50) and 1.5 AP (ResNet-101) improvements, respectively. 

\section{Conclusion}
In this paper, we make thorough improvements to the Mask RCNN based human pose estimation. The experimental results demonstrate that our refinement largely narrows the performance gap between one-stage and two-stage top-down methods, and we successfully keep the superior inference efficiency. This success mainly stems from a more suitable feature extraction strategy in RoIAlign and a better network structure for keypoint detection: (1) select the P2 layer of FPN to ensure high-resolution spatial information; (2) enlarge the person instance bounding box to prevent the loss of useful features of human pose; and (3) add global feature information through a Global Context Module to increase the receptive field for human pose estimation. Our improved model, using ResNet-50 as the backbone, achieved 66.8 AP with 1x training schedule, which is 2.8 AP higher than Mask RCNN. We hope our research and findings can inspire further study on one-stage human pose estimation methods, which can be more popular in real applications.

\begin{acks}
This work was supported by NSF of China (No. 62172222), China Postdoctoral Science Foundation (No. 2020M681609).
\end{acks}

\bibliographystyle{ACM-Reference-Format}
\bibliography{sample-base}

\appendix

\end{document}